\documentclass[10pt]{article}
\usepackage[a4paper, total={6in, 8in}]{geometry}
\usepackage[breaklinks=true]{hyperref}
\usepackage{breakcites}
\usepackage{graphicx}
\usepackage{array}
\usepackage{float}
\usepackage{multirow}
\usepackage{enumitem}
\usepackage{makecell}
\usepackage[misc,geometry]{ifsym}

\usepackage[displaymath,mathlines,running]{lineno}
\usepackage{lineno}
\usepackage{authblk}

\usepackage{amsmath}
\usepackage{amsfonts}
\usepackage{bbm}
\DeclareMathOperator*{\argmin}{arg\,min}

\usepackage{color}

\begin{document}
%
\title{What Makes an Effective Scalarising Function\\
for Multi-Objective Bayesian Optimisation?}
\author[1]{Clym Stock-Williams}
\author[2]{Tinkle Chugh}
\author[3]{Alma Rahat}
\author[4]{Wei Yu}

\affil[1]{TNO, Westerduinweg 4, 1755LE Petten, The Netherlands\newline Clym.Stock-Williams@tno.nl}
\affil[2]{Department of Computer Science, University of Exeter, UK\newline t.chugh@exeter.ac.uk}
\affil[3]{Department of Computer Science, Swansea University, UK\newline a.a.m.rahat@swansea.ac.uk}
\affil[4]{Faculty of Aerospace Engineering, Delft University of Technology, The Netherlands\newline w.yu@tudelft.nl}

\date{} 
\maketitle
%
%


\begin{abstract}
Performing multi-objective Bayesian optimisation by scalarising the objectives 
avoids computation of expensive multi-dimensional integral-based acquisition functions, instead allowing one-dimensional standard acquisition functions\textemdash such as Expected Improvement\textemdash to be applied. Here, two infill criteria based on hypervolume improvement\textemdash one recently-introduced and one novel\textemdash are compared with the multi-surrogate Expected Hypervolume Improvement. The reasons for the disparities in these methods' effectiveness in maximising the hypervolume of the acquired Pareto Front are investigated.
In addition, the effect of the surrogate model mean function on exploration and exploitation is examined: careful choice of data normalisation is shown to be preferable to the exploration parameter commonly used with the Expected Improvement acquisition function.
Finally, the effectiveness of all the methodological improvements defined here are demonstrated on a real-world problem: the optimisation of a wind turbine blade aerofoil for both aerodynamic performance and structural stiffness. With effective scalarisation, Bayesian optimisation finds a large number of new aerofoil shapes which strongly dominate standard designs.

\end{abstract}

\section{Introduction}
The class of metaheuristic optimisation algorithms have proven to be powerful at finding global optima in cases where the derivative of the optimisation objective cannot be calculated directly~\cite{Boussaid2013}. However, these algorithms usually require many thousands of evaluations to approximate an optimal solution, due to their reliance on stochastic search procedures. In many situations where optimal design is desired (for example geometry optimisation for a complex engineering component \cite{daniels:2019}) it is impractical to evaluate these many candidate designs, due to time and/or financial constraints. To alleviate this, Bayesian optimisation (BO) algorithms have been developed, first popularised for single-objective problems by Jones \textit{et al.}~\cite{Jones1998}. 

At its heart, BO learns a surrogate function of the objective space from evaluations (often with uncertainty) of candidate solutions. This surrogate is used to acquire new candidate solutions to evaluate, by explicitly maximising a trade-off between exploitation of existing high-performing areas, and exploration of previously un-explored areas~\cite{Shahriari2016}.

The surrogate is commonly a Gaussian process, conventionally constructed with a zero mean prior. In multi-objective optimisation problems there are typically two approaches to surrogate modelling. The most common is to build surrogates for each objective function (referred to here as the multi-surrogate approach)~\cite{Emmerich2016}. Another technique (here described as the mono-surrogate approach) is to build one surrogate after scalarising the multiple objectives, e.g.~\cite{Knowles2005,Rahat2017}. A recent survey comparing different scalarising functions (or `infill criteria') can be found in~\cite{Chugh_SCF_2020}. 

Mono-surrogate approaches are usually computationally faster\textemdash but under-perform\textemdash multi-surrogate approaches. This can be attributed to the fact that the scalarising functions can hide important details of the multi-objective function landscape, and thus reduce the effectiveness of the surrogate model in some problems. Here we aim to develop further insights into how to construct an effective mono-surrogate infill criterion. To do this, we consider one of the most successful infill criterion and propose a variant. We then compare performance (also against the standard multi-surrogate approach) on standard multi-objective test problems and the real-world optimisation problem of wind turbine aerofoil design.

In addition to the scalarising function, two key elements in Gaussian process surrogate modelling are the choice of mean and covariance (or kernel) functions. The choice of covariance function is the subject of active research. For example, a kernel search meta-optimisation loop has recently been shown to improve robustness to different objective function surfaces~\cite{Malkomes2018}. 
However, the literature is thin on the effect of mean functions on optimisation effectiveness; and in particular, there are none for BO when solving multi-objective problems. In this paper (taking inspiration from~\cite{Maree2020}) we first focus on this topic, primarily from a mono-surrogate BO approach. 

To summarise, the contributions of this article are as follows:

\begin{enumerate}
    \item New insights into the use of mono-surrogate approaches in Bayesian multi-objective optimisation;
    \item Examination of the effectiveness of the mean function in mono-surrogate (or scalarisation) based approaches;
    \item Demonstration of performance on standard and real-world multi-objective optimisation problems.
\end{enumerate}

The rest of this paper is organised by those key contributions, as follows:

\begin{itemize}[noitemsep]
    \item Section~\ref{sec:background} provides the necessary background to both single-objective and multi-objective BO;
    \item The role of controlling the surrogate model mean function is investigated in Section~\ref{sec:explore_exploit};
    \item Section~\ref{sec:infill-criteria} establishes the properties of two related mono-surrogate infill criteria; before
    \item Section~\ref{sec:performance-analysis} demonstrates their performance on standard test problems. Important conclusions are then drawn on what makes a good infill criterion. Following this, the approaches are applied to a real-world problem\textemdash wind turbine aerofoil design\textemdash to demonstrate that the expected performance improvements are realised.
    \item Finally, section~\ref{sec:conclusions} summarises the conclusions and highlights further opportunities for research.
\end{itemize}

\section{Bayesian Optimisation: Background and Investigation}\label{sec:background}
\subsection{Overview of the Method for Single-Objective Optimisation Problems}\label{subsec:ego-single}
We now define the BO method, for a minimisation problem. The desired outcome is to find:
\begin{linenomath*}\begin{equation*}
 \boldsymbol{x}^* = \argmin_{ \boldsymbol{x}\in \mathcal{X}} f\left( \boldsymbol{x} \right)
\end{equation*}\end{linenomath*}
where $\mathcal{X}$ is the space of feasible solutions, a constrained subset of $\mathbb{R}^D$. The search is initialised by generating $n_{init}$ candidate solutions in $\mathcal{X}$, often created by means of a Latin Hypercube~\cite{Koehler1996}. The optimisation loop then begins, commencing with the evaluation of $y_i = f(\boldsymbol{x}_i)$, where $i \in [1, n_{init}]$.

The surrogate model is first fitted to the data: usually a Gaussian process with a Mat\'{e}rn-5/2 covariance kernel $\kappa (\boldsymbol{x},\boldsymbol{\theta)}$~\cite{Snoek2012} with hyperparameters $\boldsymbol{\theta}$. This surrogate can then be used to predict function values with uncertainty at any other point $\boldsymbol{x_*}$ in the decision space:
\begin{linenomath*}\begin{align}\label{eqn:gp-prediction}
 p\left( y_* | \boldsymbol{x_*},X,\boldsymbol{y},\boldsymbol{\theta} \right) =
 \mathcal{N}\bigl( \mu(\boldsymbol{x_*}), \sigma^2(\boldsymbol{x_*})\bigr),
\end{align}\end{linenomath*}
where we can compute the mean and the variance of the Normal posterior predictive distribution using the following formulae:
\begin{linenomath*}\begin{equation}\label{eqn:gp-mean-variance}
 \begin{aligned}
  \mu(\boldsymbol{x_*}) &= \kappa(\boldsymbol{x_*}, X) \kappa(X, X) ^{-1} \boldsymbol{y},\\
  \sigma^2(\boldsymbol{x_*}) &= \kappa(\boldsymbol{x_*}, \boldsymbol{x_*}) - \kappa(\boldsymbol{x_*}, X) ^T \kappa(X,X) ^{-1} \kappa(X, \boldsymbol{x_*}).
 \end{aligned}
\end{equation}\end{linenomath*}

The next candidate solution is now selected through optimisation of an acquisition function applied to the surrogate's predictions. This acquisition function calculates the usefulness of any new point in objective space, based on the current information. One of the most common, effective and intuitive acquisition functions is the Expected Improvement (EI) over the current best-found function value $y^*$~\cite{Jones1998}:
\begin{linenomath*}\begin{align*}
 I(\boldsymbol{x}) &= max(y^* - \hat{y}, 0)\\
 \mathbb{E}[I(\boldsymbol{x})] &= \int_{-\infty}^\infty max(y^* - \hat{y}, 0) F(y) dy,
\end{align*}\end{linenomath*}
where $\hat{y}$ is the best found fitness value. If we assume the Gaussian process surrogate from equation~\ref{eqn:gp-prediction}, and substitute $y=\mu\left( \boldsymbol{x}\right) + \sigma\left( \boldsymbol{x}\right)\epsilon$ (where $\epsilon$ is the standard Normal distribution, with PDF $\phi$ and CDF $\Phi$), this gives:
\begin{linenomath*}\begin{equation}\label{eqn:expected-improvement}
 \mathbb{E}[I(\boldsymbol{x})] = \left( y^* - \mu\right) \Phi\left(\frac{y^* - \mu}{\sigma}\right) + \sigma\phi\left(\frac{y^* - \mu}{\sigma}\right) 
\end{equation}\end{linenomath*}
Optimisation of the acquisition function is then performed (using another optimisation algorithm, such as CMA-ES or L-BFGS with multiple restarts~\cite{Shahriari2016}) to select one or more locations to evaluate next.
The optimisation loop then continues until the allowed budget of evaluations is exhausted.

\subsection{Multi-objective Bayesian Optimisation}\label{subsec:ego-multi}
The goal in multi-objective optimisation is to find a set of mutually non-dominated solutions $\mathcal{P}$, as close as possible to the global Pareto Front~\cite{Emmerich2018}. The area dominated by this set, relative to a reference point $R$, is called the hypervolume ($HV$).

\begin{figure}[tb]
 \centering
 \includegraphics[width=0.5\textwidth]{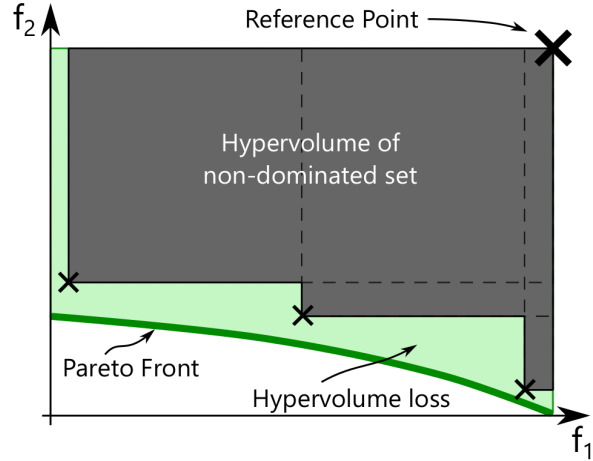}
 \caption{Schematic showing the hypervolume loss between a non-dominated set of solutions found by a multi-objective optimisation algorithm and the true Pareto Front.} \label{fig:hypervolume-loss}
\end{figure}

Maximising this hypervolume is normally performed with the same fundamental process as that described in section~\ref{subsec:ego-single}. A new acquisition function is defined, which targets minimising the loss between the global Pareto Front and the found non-dominated set, see Fig.~\ref{fig:hypervolume-loss}. The direct extension of EI to multiple objectives is Expected Hypervolume Improvement (EHVI). Recent efforts have largely focused on making the hypervolume and expected hypervolume improvement calculations faster, although usually at the expense of assuming zero uncertainty on the evaluated solutions~\cite{Fonseca2006,Beume2009,Couckuyt2014,Yang2017,Yang2019}.

\subsection{Motivation for Mono-Surrogate Infill Criteria}\label{subsec:infill-motivation}
An alternative approach to defining a new acquisition function is to use a scalarising function (or `infill criterion'). The reasons to choose this approach are:
\begin{enumerate}
 \item the existing acquisition functions and software for single-objective optimisation can be used;
 \item the expensive process of fitting multiple surrogate models can be avoided;
 \item the very large number of expensive integrations required for EHVI and similar acquisition functions can be avoided.
\end{enumerate}

The simplest version is to calculate a weighted mean of the objective function values for each solution, and then apply EI. This, however, has the unfortunate side-effect of directing the optimisation down a single vector in the objective space. 

More sophisticated approaches have been developed in the BO literature~\cite{Rahat2017} and it is now known that Pareto-compliance is a pre-requisite for effectiveness~\cite{Zitzler2007}. Nevertheless, this leaves open a large design space for such functions, and no principled approach to increase effectiveness is apparent in the previous literature proposing new scalarisation functions. Further, analyses of their effectiveness have been largely limited to empirical studies and plots of their theoretical properties in objective space.

\section{Effect of Mean Function in Balancing Exploration and Exploitation}\label{sec:explore_exploit}

To the knowledge of the authors, the balance of exploitation versus exploration has previously focused entirely on the acquisition function~\cite{Ath2019,Zhan2020}, neglecting the important role the prior mean takes in determining the predictive mean and variance.

Despite being poorly-reviewed in the literature, a `jitter' parameter $\zeta$ is commonly included in equation~(\ref{eqn:expected-improvement})~\cite{Berk2018}:
\begin{linenomath*}\begin{equation}\label{eqn:expected-improvement-with-jitter}
 \mathbb{E}[I(\boldsymbol{x})] = \left( y^* + \zeta - \mu\right) \Phi\left(\frac{y^* + \zeta - \mu}{\sigma}\right) + \sigma\phi\left(\frac{y^* + \zeta - \mu}{\sigma}\right) 
\end{equation}\end{linenomath*}

The intention is clearly to enhance exploration by making the current best function evaluation slightly worse. The default value used in the Python package GPyOpt, for example, is $\zeta = 0.01$~\cite{gpyopt2016}.

\begin{figure}[H]
 \includegraphics[width=\textwidth]{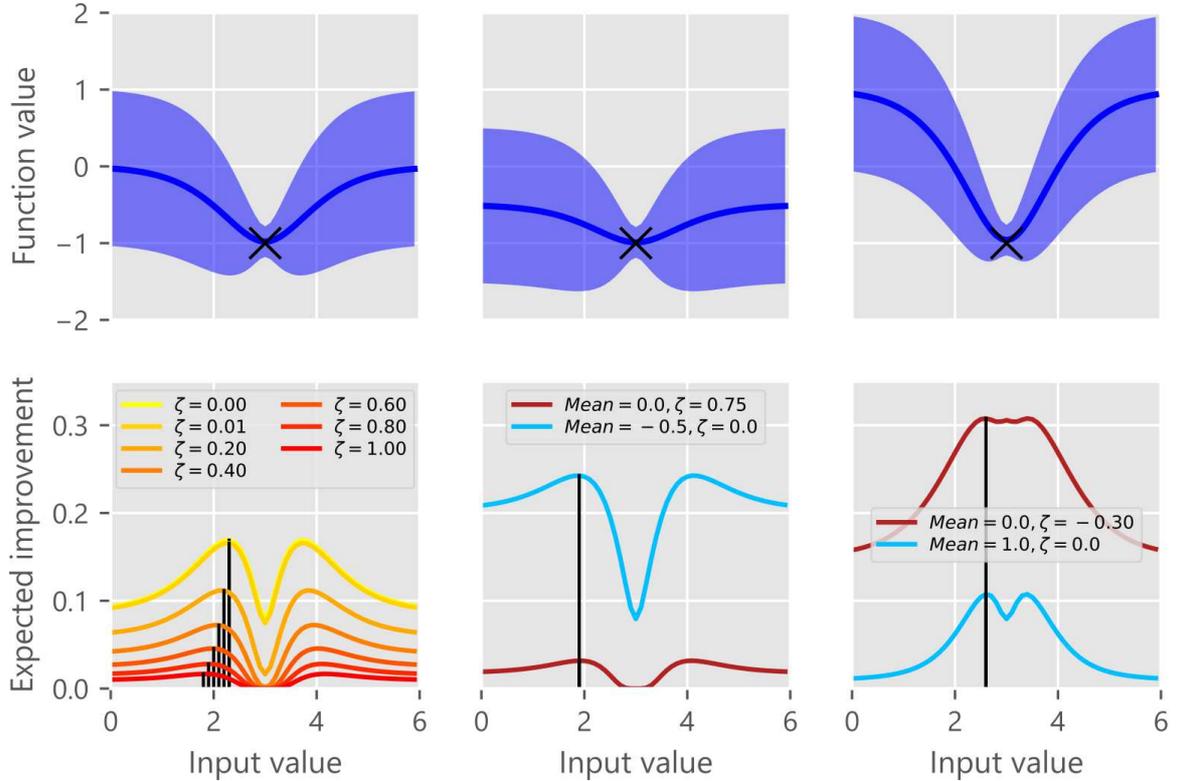}
 \caption{Comparison of the effect of the mean function and jitter parameter $\zeta$ on the exploration of the Expected Improvement (EI) acquisition function. Top row: Gaussian process surrogate with varying mean and constant kernel variance and lengthscale, provided evaluated solutions marked with a black `x'. Bottom row: EI calculations. Left: when the prior mean is 1 standard deviation above the evaluated solution, the change with $\zeta$ of the optimum EI location is shown. Centre: when the mean is closer to the evaluated solution, more exploration is performed. Right: when the mean is worse, more exploitation is performed. For both centre and right lower plots, the red line shows the $\zeta$ parameter value required to obtain the same EI optimum when using the zero mean function value from the left plot.} \label{fig:acquisition-exploration}
\end{figure}

We now compare the effects of changing this $\zeta$ parameter with varying the prior mean relative to the current best. 
As shown in Fig.~\ref{fig:acquisition-exploration}, comparatively very large values of $\zeta$ are required to significantly change the exploration behaviour of the acquisition function. 
On the other hand, the relationship between the surrogate mean function value and the shape of the EI surface is intuitive and better-behaved. As the mean improves, the value already found becomes less surprising, thus exploration is preferred. Conversely, worsening the mean increases the surprise of the value found, making exploitation preferable. 
Note further the poor shape of the EI function surface as $\zeta$ becomes negative in the lower right plot of Fig.~\ref{fig:acquisition-exploration}.

Returning to equations~(\ref{eqn:gp-prediction}),~(\ref{eqn:gp-mean-variance}),~(\ref{eqn:expected-improvement}) and~(\ref{eqn:expected-improvement-with-jitter}), the equivalence between $\zeta$ and a mean function increase of value $\lambda$ can quickly be derived, for the situation in which only one solution has been acquired:
\begin{linenomath*}\begin{equation*}
 y^* + \zeta - \kappa\left( \boldsymbol{x_*}, \boldsymbol{x}\right) \kappa\left( \boldsymbol{x}, \boldsymbol{x}\right) ^{-1} y^* = 
 \left( y^* + \lambda\right) - 
 \kappa\left( \boldsymbol{x_*}, \boldsymbol{x}\right) \kappa\left( \boldsymbol{x}, \boldsymbol{x}\right) ^{-1} \left( y^* + \lambda\right)
\end{equation*}\end{linenomath*}
\begin{linenomath*}\begin{equation}\label{eqn:expected-improvement-with-jitter2}
 \zeta = \lambda\left( 1 - \kappa\left( \boldsymbol{x_*}, \boldsymbol{x}\right) \kappa\left( \boldsymbol{x}, \boldsymbol{x}\right) ^{-1}\right)
\end{equation}\end{linenomath*}

The jitter method therefore does not apply a constant offset to the prior mean, but instead a less-intuitive offset that changes with distance in the decision space from the existing solutions. 

Instead of altering the acquisition function, it is therefore recommended to focus on carefully choosing the prior mean (or, equivalently, normalising the measured function values for a zero-mean prior). 
The following approach is therefore suggested (given the usual situation where the global minimum function value of an optimisation objective is unknown). Each time a new (batch of) candidate solutions are desired, the archive of solution values is renormalised using equation~\ref{eqn:normalising-mean} before fitting the zero-mean Gaussian process surrogate.
\begin{linenomath*}\begin{equation}\label{eqn:normalising-mean}
 y\prime = \left( 1 - \xi\right) \cdot \overline{y} + \xi \cdot y^*
\end{equation}\end{linenomath*}

Setting the exploration parameter $\xi = 0.0$ implies using the current mean $\overline{y}$ as the prior mean. Positive values promote more exploratory behaviour, while negative values promote exploitation. In the following sections, this approach will be applied.

\section{Extending Hypervolume Improvement}\label{sec:infill-criteria}
\subsection{Defining Mono-Surrogate Infill Criteria}\label{subsec:xhvi-definition}
We now define a novel infill criterion which fulfills the criteria from section~\ref{subsec:infill-motivation}. Instead of estimating the hypervolume improvement for predicted objective function values, we shall instead follow the approach in~\cite{Ponweiser2008} and calculate the hypervolume improvement contributed by each existing solution evaluated with expensive objective functions. We store all the evaluated solutions in an Archive ($\mathcal{A}$).

\begin{figure}[bth]
 \includegraphics[width=\textwidth]{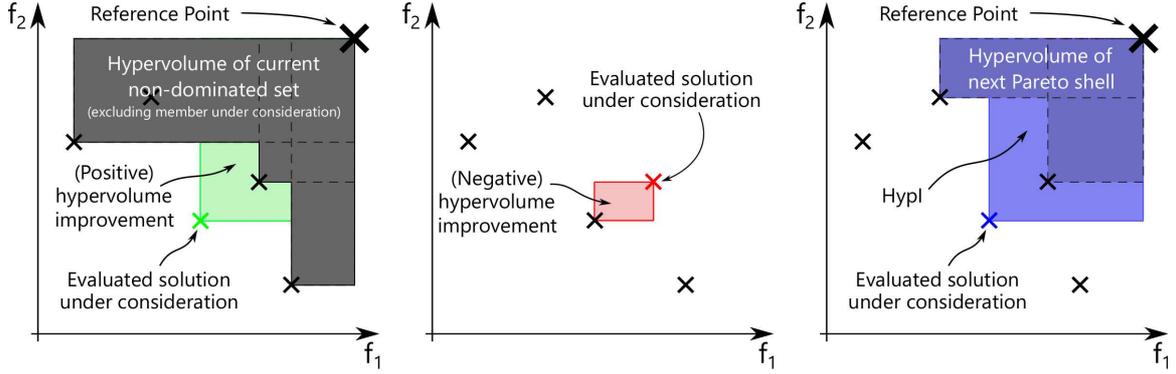}
 \caption{Schematic showing the calculation of Extended Hypervolume Improvement (xHVI) and HypI. Left: positive hypervolume improvement for the non-dominated solution marked in green; centre: negative hypervolume improvement for the dominated solution marked in red; right: HypI for the solution marked in blue.} \label{fig:hypervolume-improvement}
\end{figure}

An illustration for a bi-objective minimisation problem is shown in Fig.~\ref{fig:hypervolume-improvement}. For a non-dominated solution\textemdash see the left-hand schematic in Fig.~\ref{fig:hypervolume-improvement}\textemdash this shall be a positive number:
\begin{linenomath*}\begin{equation}\label{eqn:hvi-plus}
 \mathrm{HVI^+}(\boldsymbol{x},R) = \mathrm{HV}(\mathcal{A} \setminus \{\boldsymbol{x}\},R) - \mathrm{HV}(\mathcal{A}, R).
\end{equation}\end{linenomath*}
Since we want to maximise the total hypervolume of the non-dominated set, this metric gives an intuitive measure of the contribution of each non-dominated solution, as well as penalising solutions which are close together in the objective space.

However, a zero value is assigned to dominated solutions, whereas we want a smoothly varying gradient over the decision space. We therefore estimate how poor dominated solutions are by calculating a `negative hypervolume improvement', as shown in the centre schematic of Fig.~\ref{fig:hypervolume-improvement}:
\begin{linenomath*}\begin{equation}\label{eqn:hvi-minus}
 \mathrm{HVI^-}(\boldsymbol{x}) = \mathrm{HV}(\mathcal{P}, \boldsymbol{x})
\end{equation}\end{linenomath*}
 
This naturally assigns non-dominated solutions a zero value. Thus the infill criterion, Extended Hypervolume Improvement (xHVI), is defined as follows:
\begin{linenomath*}\begin{equation}\label{eqn:xhvi}
 \mathrm{xHVI}(\boldsymbol{x};R) = \mathrm{HVI^+}(\boldsymbol{x};R) - \mathrm{HVI^-}(\boldsymbol{x})
\end{equation}\end{linenomath*}

We shall compare the performance of this infill criterion with the recently-proposed HypI~\cite{Rahat2017}, shown in the right-hand schematic of Fig.~\ref{fig:hypervolume-improvement}. The definition of this requires that we assign each solution to a Pareto shell $\mathcal{P}_k$, where the second Pareto shell $\mathcal{P}_2$ is the set of mutually non-dominated solutions once all members of $\mathcal{P}_1 = \mathcal{P}$ are removed from the archive, and so on. Each solution is assigned the following value:
\begin{linenomath*}\begin{equation}\label{eqn:hypi}
 \mathrm{HypI}(\boldsymbol{x},R) =  \mathrm{HV}(\mathcal{P}_{k+1} \cup \{\boldsymbol{x}\}, R)
\end{equation}\end{linenomath*}

Conceptually, this is rather similar to xHVI. A solution's value increases with the amount of hypervolume it generates. The main difference apparent at this stage is that it does not penalise solutions by their proximity in objective space (note that this is performed by the acquisition function).

\subsection{Illustration on bi-objective optimisation problem}\label{subsec:xhvi_use}
\begin{figure}[!htb]
 \includegraphics[width=\textwidth]{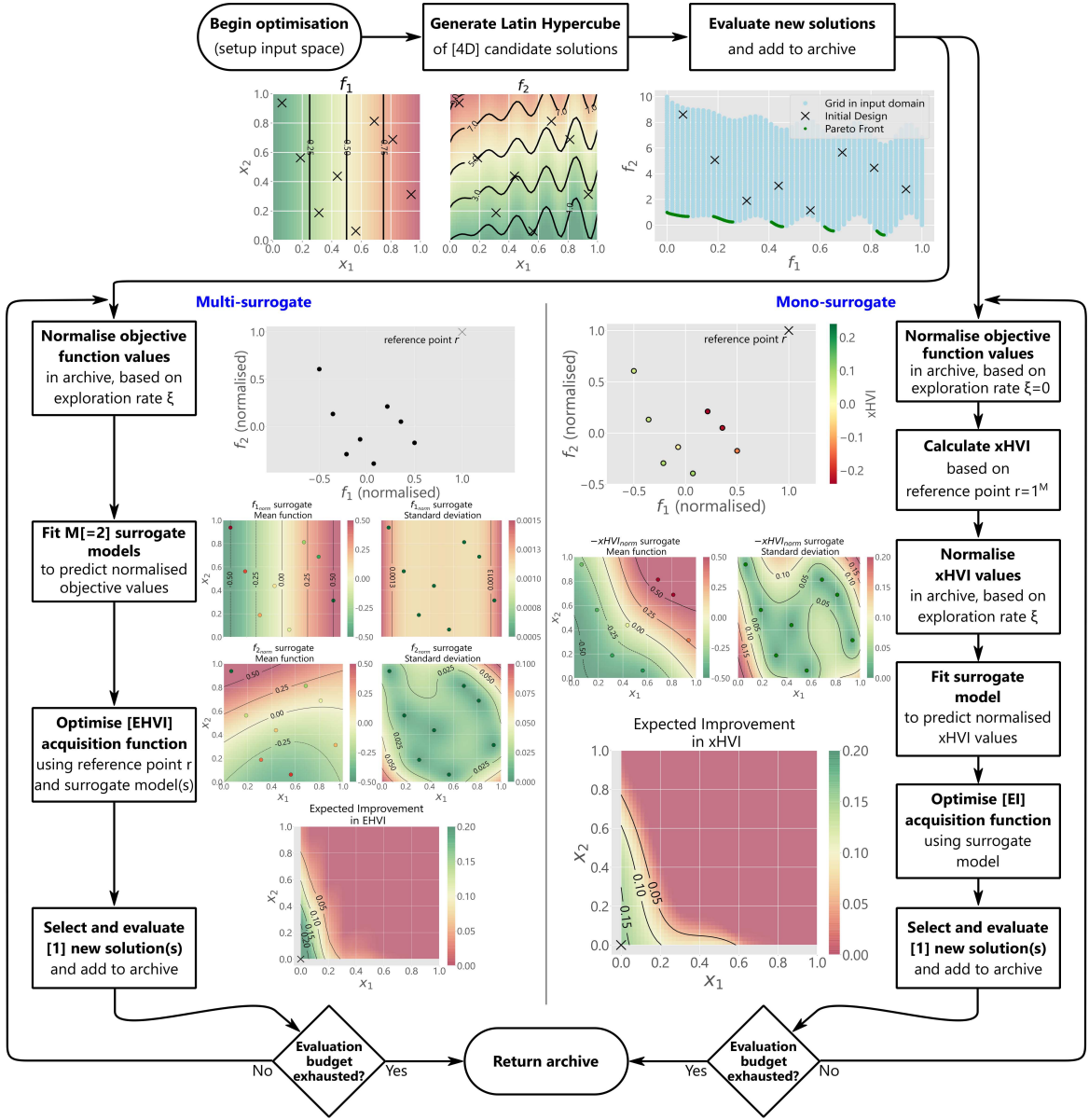}
 \caption{Process diagram comparing multi-objective Bayesian optimisation with (right hand side) and without (left hand side) the use of a mono-objective infill criterion. The illustrative examples shown use the ZDT3 test function with 2 input dimensions and 2 objectives. Values in square brackets refer to the choices made in section~\ref{subsec:performance}.} \label{fig:optimisation-process}
\end{figure}

The optimisation process outlined in section~\ref{subsec:ego-single} is now elaborated in Fig.~\ref{fig:optimisation-process} for both multi- and mono-surrogate BO, illustrating similarities and differences using the ZDT3 test function~\cite{Zitzler2000} with $D=2$ input dimensions and $M=2$ objectives.

Initially, as shown on the top row of Fig.~\ref{fig:optimisation-process}, a space-filling design is created. The exact balance of benefit in the proportion of the evaluation budget used up here and in the optimisation will depend on the properties of the function and the optimisation method chosen. With EHVI, for example, it is important to obtain a good initial surrogate fit to the objective functions. A Latin Hypercube of size $4D=8$ is used here, immediately finding two solutions reasonably close to the Pareto Front. A grid of 2500 solutions in the decision space is provided (in light blue) for reference.

With the standard BO process described in section~\ref{subsec:ego-single}, we proceed down the left hand side of the process diagram. 
The objective function values are first normalised as described in section~\ref{sec:explore_exploit}, using $\xi=0.0$ (i.e. the mean of each set of evaluated objective values is zero). 
Two independent surrogate models are fitted, one for each objective. As can be seen (referring back to the plots at the top of the figure for the true objective surfaces) the independence of $f_2$ with $x_2$ is captured, and the general shape of $f_2$ is also realised. 

Finally, the Expected Hypervolume Improvement is calculated\textemdash the surface is shown in the lowest plot\textemdash  using the reference point $r=\mathbbm{1}^M$ and 1000 Monte Carlo samples. 
The acquisition function is optimised with L-BFGS, with empirical gradients, conducting 10 re-starts from initial locations which are chosen randomly close to the existing non-dominated set. The next solution to acquire is $\left[ 0.0, 0.0\right]$, which is immediately on the true Pareto Front.

Now, we turn our attention to the right hand side of Fig.~\ref{fig:optimisation-process}, where the xHVI infill criterion is used. 
The first step is to calculate the xHVI for each evaluated solution. The top plot illustrates how the objective values are first normalised as for the multi-surrogate process. 
Then equations~\ref{eqn:hvi-plus}-\ref{eqn:xhvi} are applied, based on a reference point $r=\mathbbm{1}^M$. This reference point can also be fixed from knowledge of the lowest useful objective value.

Once this is achieved, only one surrogate is fitted. As can be seen in the middle plots, the shape of the surrogate fits well the shape of the non-dominated set. Optimisation of Expected Improvement (again with  L-BFGS, with empirical gradients) can proceed as usual for single-objective BO. 
The next solution to acquire is in agreement with EHVI: $\left[ 0.0, 0.0\right]$.

\section{Performance Analysis on Multi-Objective Problems}\label{sec:performance-analysis}
\subsection{Performance on Standard Multi-objective Test Functions}\label{subsec:performance}
The methodologies described in section~\ref{subsec:xhvi_use} for EHVI-, xHVI- and HypI-based BO are now applied to functions from the well-known ZDT~\cite{Zitzler2000} and DTLZ~\cite{Deb2002} suites, as implemented in the DEAP Python toolbox~\cite{DEAP_JMLR2012}. $D = 10$ input dimensions and $M = 2$ objectives are used for all functions.

The evaluation budget is restricted to 300, which is reasonable for the expensive problems encountered in practice. A Latin Hypercube design of 40 candidates is provided, using the PyDOE Python toolbox. Thus 260 additional evaluations are available for BO, which is run sequentially.

Two free parameters have been identified for the optimisation processes:
\begin{enumerate}
 \item The exploration parameter $\xi$ from section~\ref{sec:explore_exploit}; and
 \item The value of the normalised reference point $r$ from section~\ref{subsec:xhvi_use}.
\end{enumerate}
It is likely that these hyperparameters can be tuned for every problem, however, one choice will now be made for all functions.
First, $\xi$ is set to $0.0$. As discussed in~\ref{subsec:xhvi_use}, this corresponds to the commonly-assumed case where the evaluated objective function values are normalised to their mean.
Second, the reference point $r$ is set to $\mathbbm{1}^M$.

Each optimisation is run 21 times, each iteration using a different initial random seed for the Latin Hypercube, in order to provide statistics on the outcomes of the optimisation. The same random seeds are used to initialise the xHVI, HypI and EHVI runs, to ensure a fair test of the optimisation algorithms' performance.

Many performance metrics have been proposed in literature~\cite{Audet2018}, which analyse the convergence to, and spread along, the true Pareto Front achieved by the final non-dominated set, and the efficiency of the algorithm. We used the following performance metrics in this work:
\begin{itemize}
    \item \textbf{Hypervolume (HV) or S-metric} - presented here as a percentage of the achievable hypervolume (see also Fig.~\ref{fig:hypervolume-loss}). The reference point required for its calculation is chosen to be 10\% greater than the true Pareto Front's nadir point.
    \item \textbf{Non-dominated set size} - this indicates the number of meaningful choices presented to the user of the optimisation.
    \item \textbf{Empirical attainment function (EAF)} - proposed by~\cite{Fonseca1996}, this is the empirical cumulative distribution over the probability of a point in objective space being dominated by the obtained non-dominated set.
\end{itemize}

\begin{figure}[tbh]
 \centering
 \includegraphics[width=0.75\textwidth]{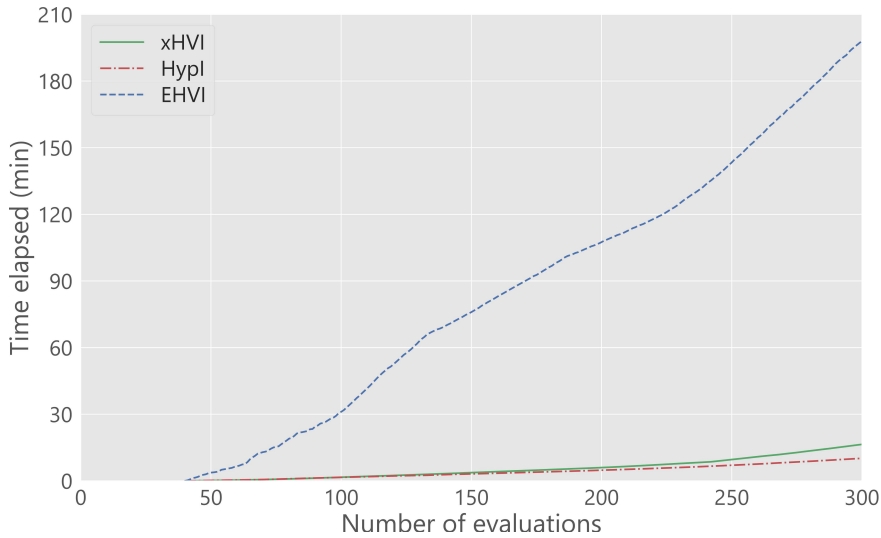}
 \caption{Comparison of computation times required for xHVI, HypI and EHVI Bayesian optimisation methods (using ZDT6) on a 4-core i7-6700HQ CPU.} \label{fig:comp-time}
\end{figure}

Since the evaluation budget is very limited, little attention is given to efficiency. Comparison of runtime on one problem for xHVI and EHVI is, however, provided for interest in Fig.~\ref{fig:comp-time}. 

\subsection{Analysis}
Beginning with Fig.~\ref{fig:comp-time}, it is apparent that the runtime of EHVI\textemdash due the need to fit twice as many surrogate models, and conduct extensive Monte Carlo calculations of HVI to optimise the acquisition function\textemdash is more than 2.5 times that of xHVI or HypI. With an expensive evaluation function and batch optimisation this difference will become less significant. 

\begin{figure}[H]
 \centering
 \includegraphics[width=\textwidth,height=0.85\textheight,keepaspectratio]{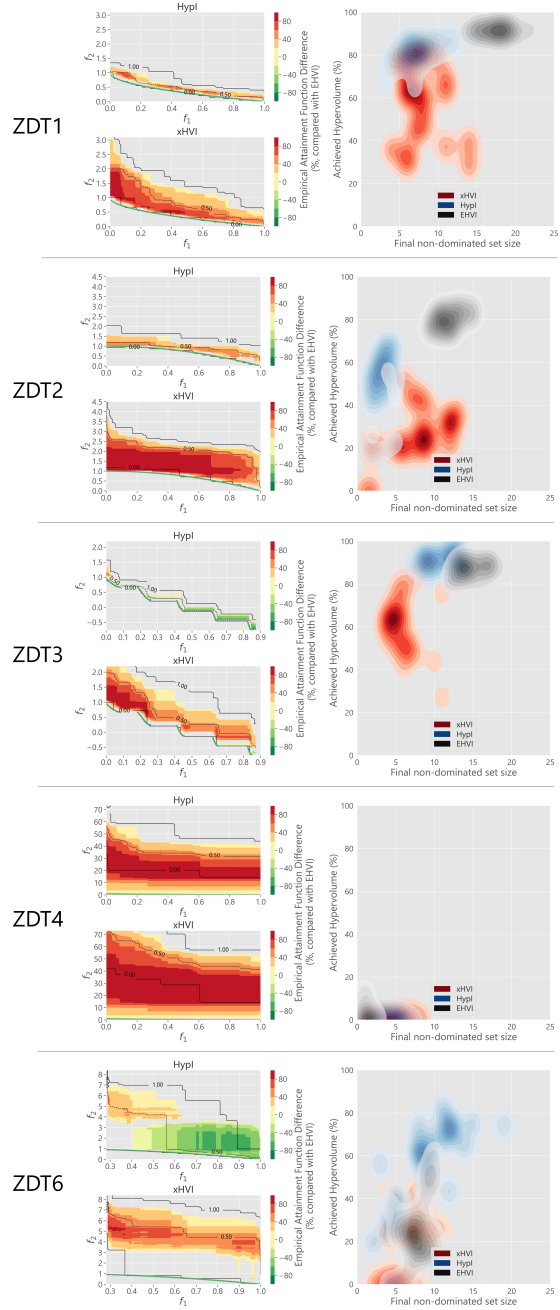}
 \caption{Comparison of 21 optimisations with 300 evaluations using the EHVI, xHVI and HypI Bayesian optimisation approaches, on five ZDT test functions each with 10 dimensions and 2 objectives. Left: Empirical attainment functions (EAFs) for the final non-dominated sets (black lines), the difference with the EAF of EHVI (colour), and the true Pareto Front (green). Right: Bivariate kernel density estimates of the \% hypervolume achieved and non-dominated set size, for all three approaches.} \label{fig:results-ZDT}
\end{figure}

\begin{figure}[H]
 \centering
 \includegraphics[width=\textwidth,height=0.85\textheight,keepaspectratio]{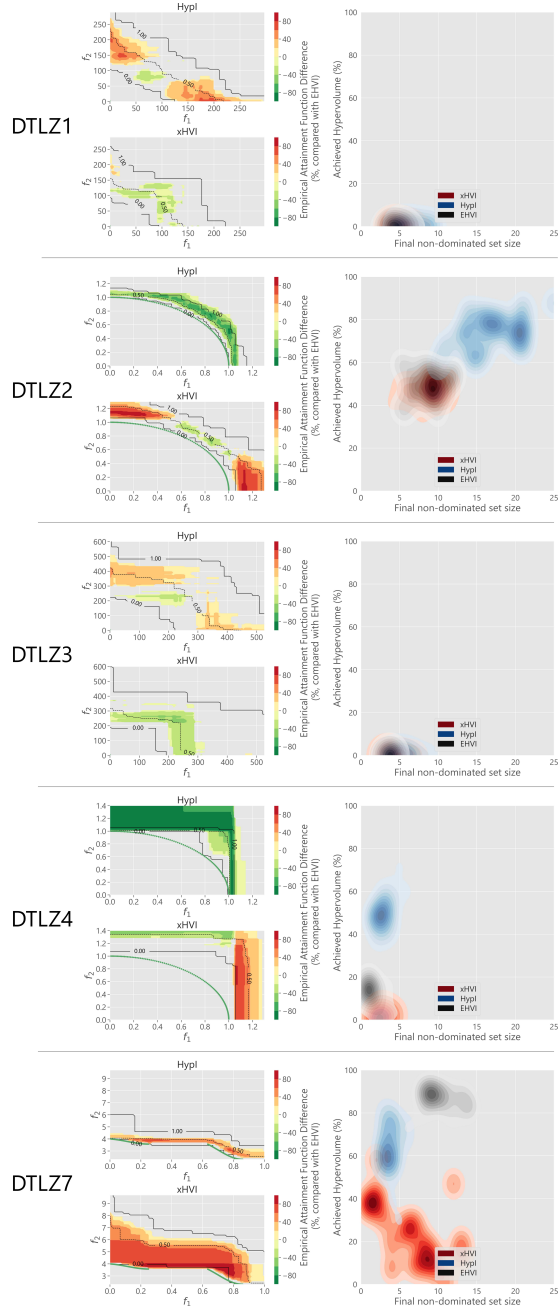}
 \caption{Comparison of 21 optimisations with 300 evaluations using the EHVI, xHVI and HypI Bayesian optimisation approaches, on five DTLZ test functions each with 10 dimensions and 2 objectives. Left: Empirical attainment functions (EAFs) for the final non-dominated sets (black lines), the difference with the EAF of EHVI (colour), and the true Pareto Front (green). Right: Bivariate kernel density estimates of the \% hypervolume achieved and non-dominated set size, for all three approaches.} \label{fig:results-DTLZ}
\end{figure}

The EAFs of HypI and xHVI are shown in Fig.~\ref{fig:results-ZDT} for ZDT and Fig.~\ref{fig:results-DTLZ} for DTLZ problems. We also show the differences in EAF of HypI and xHVI in comparison to EHVI. Next to the EAFs, we present bivariate kernel density plots of (\%) hypervolume and non-dominated set size. This information is now discussed and analysed.

It is clear from these figures that EHVI and HypI demonstrate impressive performance in reliably converging to the true Pareto Front on ZDT1, ZDT2, ZDT3, DTLZ2 and DTLZ7. Conversely, none of the algorithms are able to solve ZDT4, DTLZ1 or DTLZ3 adequately.

The colours in the left-hand objective space plots (and the vertical axis of the right-hand graphs) give an impression of the relative performance of EHVI and the mono-objective methods. To quantify these differences, it is also informative to (double-)integrate over the EAF differences to obtain a single numerical estimate of the relative performance of three methods; the results are shown in Table~\ref{tab:eaf_integral_results}.

\begin{table}[th]
 \centering
 \begin{tabular}{c|c|c}
  \hline
  \thead{Test function} & \thead{Integrated EAF difference\\HypI-EHVI} & \thead{Integrated EAF difference\\xHVI-EHVI} \\
  \hline\hline
   ZDT1 & $-0.09$ & $-0.50$\\
   ZDT2 & $-0.19$ & $-1.2$\\
   ZDT3 & $0.04$ & $-0.42$\\
   ZDT4 & $-25$ & $-41$\\
   ZDT6 & $0.50$ & $-1.2$\\
   DTLZ1 & $-5.5\times10^3$ & $2.0\times10^3$\\
   DTLZ2 & $0.11$ & $-0.13$\\
   DTLZ3 & $-1.7\times10^4$ & $1.5\times10^4$\\
   DTLZ4 & $0.62$ & $0.02$\\
   DTLZ7 & $-0.27$ & $-1.6$\\
   \hline
 \end{tabular}
 \caption{Numerical analysis of results shown in Figs.~\ref{fig:results-ZDT} and~\ref{fig:results-DTLZ}. The double integral over the difference in empirical attainment function (EAF) is calculated, comparing each scalarising function (HypI and xHVI) with the multi-surrogate EHVI. Negative numbers indicate a worse performance than EHVI overall; positive numbers the opposite.}
 \label{tab:eaf_integral_results}
\end{table}



The performance of an optimisation algorithm can itself be viewed as a multi-objective decision making problem. Turning to the kernel density estimate plots, two metrics for assessing performance are presented, in a manner such that dominance of one optimisation algorithm over another can easily be seen. 

Maximising HV is of course the main measure of optimisation effectiveness. This metric has theoretically superior behaviour to others with which it shares a high correlation, such as inverted generational distance~\cite{Veldhuizen1998}. Any optimisation methods producing equal HV, however, can be further distinguished by the non-dominated set size. This metric\textemdash despite not distinguishing solutions which are close together in objective space\textemdash does provide a useful indicator of the amount of information available to the user of the optimisation results.

\begin{figure}[H]
 \centering
 \includegraphics[width=0.75\textwidth]{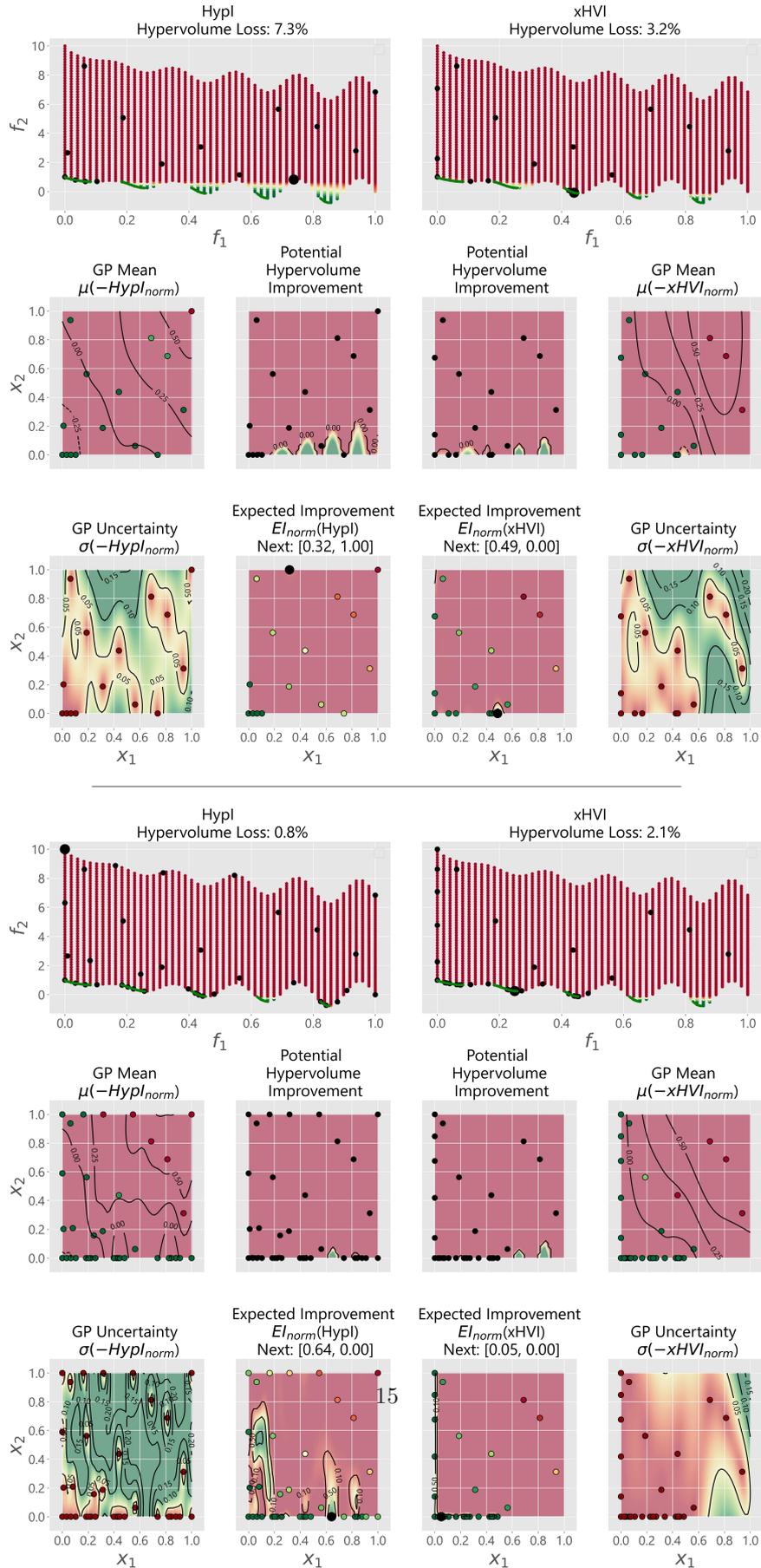}
 \caption{Illustration of optimisation for the xHVI and HypI infill criteria, using the ZDT3 test function with 2 input dimensions and 2 objectives. Top: after 15 acquired solutions; bottom: after 36 acquired solutions.}
 \label{fig:hypi_xhvi_optimisation}
\end{figure}

Comparing the performance of the algorithms themselves, xHVI is clearly a much less capable scalarising function than HypI. Determining the reasons for this requires us to return to the example from Fig.~\ref{fig:optimisation-process}, and compare their surrogate models and acquisition surfaces, which are shown for two states during the optimisation process in Fig.~\ref{fig:hypi_xhvi_optimisation}.

Each graph shows the currently acquired points, initially in the objective space, and then in decision space. ``Potential Hypervolume Improvement" indicates the best possible locations for the optimisation algorithms to sample next. By comparison, the ``Expected Improvement" graph beneath indicates the surface which is being optimised over to select the next point for acquisition (shown as a large dot). The mean and standard deviation of the Gaussian process complete the picture. They are coloured so that the two components of Equation~\ref{eqn:expected-improvement} are shown equally, and green indicates that a particular location is better for acquisition.

The key driver behind the different behaviour of the two scalarising functions is their gradients. The gradient of HypI is much more consistent over the whole surface, whereas xHVI assigns much larger values of $HVI^-$ to dominated solutions than the $HVI^+$ assigned to non-dominated solutions. 

In the early stages, this works to allow xHVI to obtain a higher hypervolume. The two sections of the Pareto Front with lowest $f_1$ are found and exploited. However, the very large negative values assigned by xHVI to the locations with higher $f_1$ conspire to force xHVI never to explore that area, despite the high uncertainty. As a result, within another 20 acquisition steps HypI has obtained a much higher hypervolume, driven by the smaller GP lengthscales in $x_1$, which allow the periodic variation in Pareto Front in that dimension to be found.

\subsection{Application to Wind Turbine Aerofoil Design}\label{sec:aerofoil}

Optimisation of aerodynamic shapes is a common topic in the literature \cite{SKINNER2018933}. Wind turbine aerofoil design under uncertainty~\cite{Caboni2018} and considering multiple objectives~\cite{Ferreira2015,DeTavernier2019} are also recent topics, using evolutionary computation. There has also been some attention to single-objective Bayesian optimisation of aerofoil shapes~\cite{Jasrasaria2018}. However, to the knowledge of the authors, there have been no Bayesian optimisation studies of aerofoils considering multiple objectives, particularly for wind turbines.

\subsubsection{Aerofoil Geometry Definition}
Several parameterisations of aerofoil shape have been used in the design~\cite{Sripawadkul2010,Masters2015}, some of which have serious limitations for optimisation, since constraints on their decision spaces can be difficult to set \textit{a priori} such that realistic aerofoils are produced.

\begin{figure}[H]
 \includegraphics[width=\textwidth]{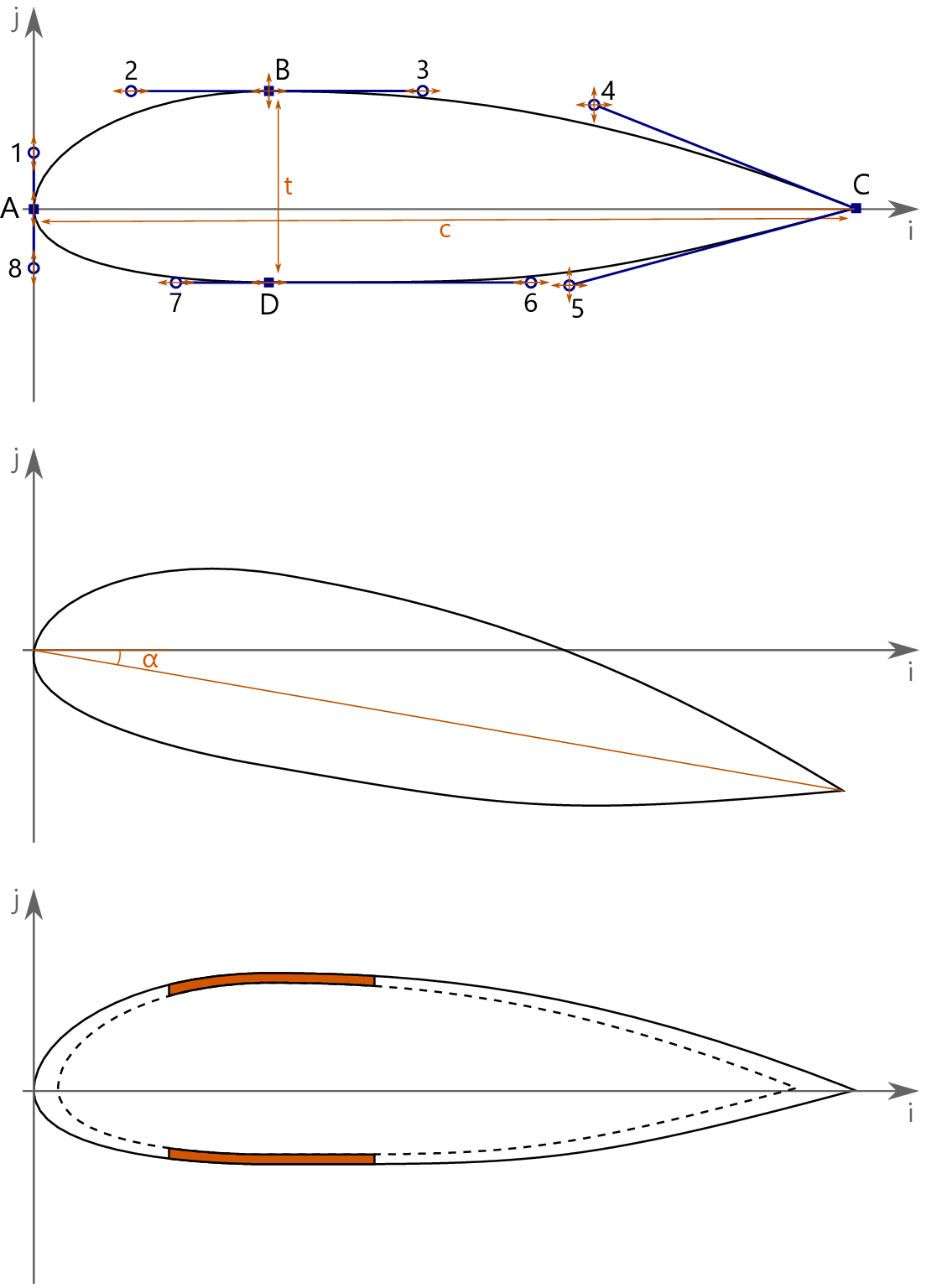}
 \caption{Top: Definition of a wind turbine blade aerofoil, using B\'{e}zier curves. The black curve is the aerofoil shape; the blue control points define the B\'{e}zier curves; and the orange arrows indicate optimisation parameters. Middle: Definition of the angle of attack $\alpha$ for aerodynamic evaluation. Bottom: Construction of a spar cap for evaluation of structural stiffness.} \label{fig:aerofoil_shape}
\end{figure}

B\'{e}zier curves offer a highly controllable\textemdash while also simple\textemdash way of defining shapes, which can be made flexible to the limitations the designer wishes to impose.
Referring to Fig.~\ref{fig:aerofoil_shape}, we choose here to optimise an aerofoil with a fixed chord $c$ and thickness $t=0.18c$ (this is a realistic constraint given that an aerofoil is a 2D cross-section through a blade with continuously changing shape). Points A (the leading edge) and C (the trailing edge) are fixed, leaving the 13-dimensional decision space given in Table~\ref{tab:aerofoil_inputs}. 

\begin{table}[th]
    \centering
    \begin{tabular}{c|l|c}
    \hline
    \thead{Decision variable} & \thead{Description} & \thead{Bounds}\\
    \hline\hline
        $j_1$ & \makecell{the vertical position of control point 1,\\\hspace{3mm}relative to A and as a fraction of $j_B - j_A$} & $[0.2,0.9]$\\
        $j_B$ & \makecell{the vertical position of B,\\\hspace{3mm}relative to A and as a fraction of t} & $[0.50,0.95]$\\
        $i_B$ & \makecell{the horizontal position of B,\\\hspace{3mm}relative to A and as a fraction of c} & $[0.15,0.70]$\\
        $i_2$ & \makecell{the horizontal position of control point 2,\\\hspace{3mm}relative to B and as a fraction of $i_B - i_A$} & $[0.1,0.9]$\\
        $i_3$ & \makecell{the horizontal position of control point 3,\\\hspace{3mm}relative to B and as a fraction of $i_C - i_B$} & $[0.2,0.8]$\\
        $i_4$ & \makecell{the horizontal position of control point 4,\\\hspace{3mm}relative to C and as a fraction of $i_C - i_3$} & $[0.05,0.75]$\\
        $j_4$ & \makecell{the vertical position of control point 4,\\\hspace{3mm}relative to C and as a fraction of $j_B - j_C$} & $[0.05,0.95]$\\
        $i_5$ & \makecell{the horijontal position of control point 5,\\\hspace{3mm}relative to C and as a fraction of $i_4 - i_6$} & $[0.05,0.75]$\\
        $j_5$ & \makecell{the vertical position of control point 5,\\\hspace{3mm}relative to C and as a fraction of $j_4 - j_D$} & $[0.05,0.50]$\\
        $i_D$ & \makecell{the horizontal position of D,\\\hspace{3mm}relative to A and as a fraction of c} & $[0.15,0.70]$\\
        $i_6$ & \makecell{the horizontal position of control point 6,\\\hspace{3mm}relative to B and as a fraction of $i_C - i_D$} & $[0.2,0.8]$\\
        $i_7$ & \makecell{the horizontal position of control point 7,\\\hspace{3mm}relative to B and as a fraction of $i_D - i_A$} & $[0.1,0.9]$\\
        $j_8$ & \makecell{the vertical position of control point 8,\\\hspace{3mm}relative to A and as a fraction of $j_A - j_D$} & $[0.2,0.9]$\\
    \hline
    \end{tabular}
    \caption{Decision space for optimising the aerofoil shown in Fig.~\ref{fig:aerofoil_shape}}
    \label{tab:aerofoil_inputs}
\end{table}

\subsubsection{Optimisation Objectives}

\paragraph{Aerodynamic performance}
A wind turbine blade, of which the aerofoil is a cross-section, should maximise its extraction of energy from the wind. It is therefore logical to maximize also the aerofoil's local contribution to the power coefficient, which is equivalent to maximising the lift:drag ratio $\frac{c_l}{c_d}$~\cite{Bjorck1988}. 

Stability of performance with variable angles of attack (the relative direction of the wind to the aerofoil axis, as shown in Fig.~\ref{fig:aerofoil_shape}) is also important. Thus the aerodynamic performance objective used here is to maximize the integral of $\frac{c_l}{c_d}$ over a range of positive angles of attack $\alpha$:

\begin{linenomath*}\begin{equation}
f_{\alpha}=\int_{\alpha=0}^{\alpha=10} \frac{c_l}{c_d} d\alpha
\end{equation}\end{linenomath*} 

The aerofoils are here evaluated using Rfoil~\cite{Rfoil} (with Mach number 0, Reynolds number $6 \times 10^{6}$ and $N_{crit} = 9$). 

\paragraph{Structural stiffness} 
To create a survivable wind turbine blade, it is important that the aerofoil can withstand high loads, while keeping manufacturing costs low. In order to evaluate the structural performance realistically\textemdash while avoiding excessive computational cost\textemdash a simple spar cap is assumed, as illustrated in Fig.~\ref{fig:aerofoil_shape}. The spar cap width is chosen as $0.25c$, with thickness calculated by scaling the aerofoil shape by 90\% and locating the centre of the spar cap at the same chord position as the upper crest (point B). 

Structural performance is evaluated by calculating the bending stiffness of the spar cap in the flapwise direction, as follows:

\begin{linenomath*}\begin{equation}
f_s= \iint_{A_{s}} E(i,j)j^2  di dj
\end{equation}\end{linenomath*}

where $A_s$ is the area of the spar cap, and the elastic modulus E is assumed to be constant at 1 GPa. 

\subsubsection{Results and Comparison with Standard Aerofoils}
\begin{figure}[H]
 \includegraphics[width=\textwidth]{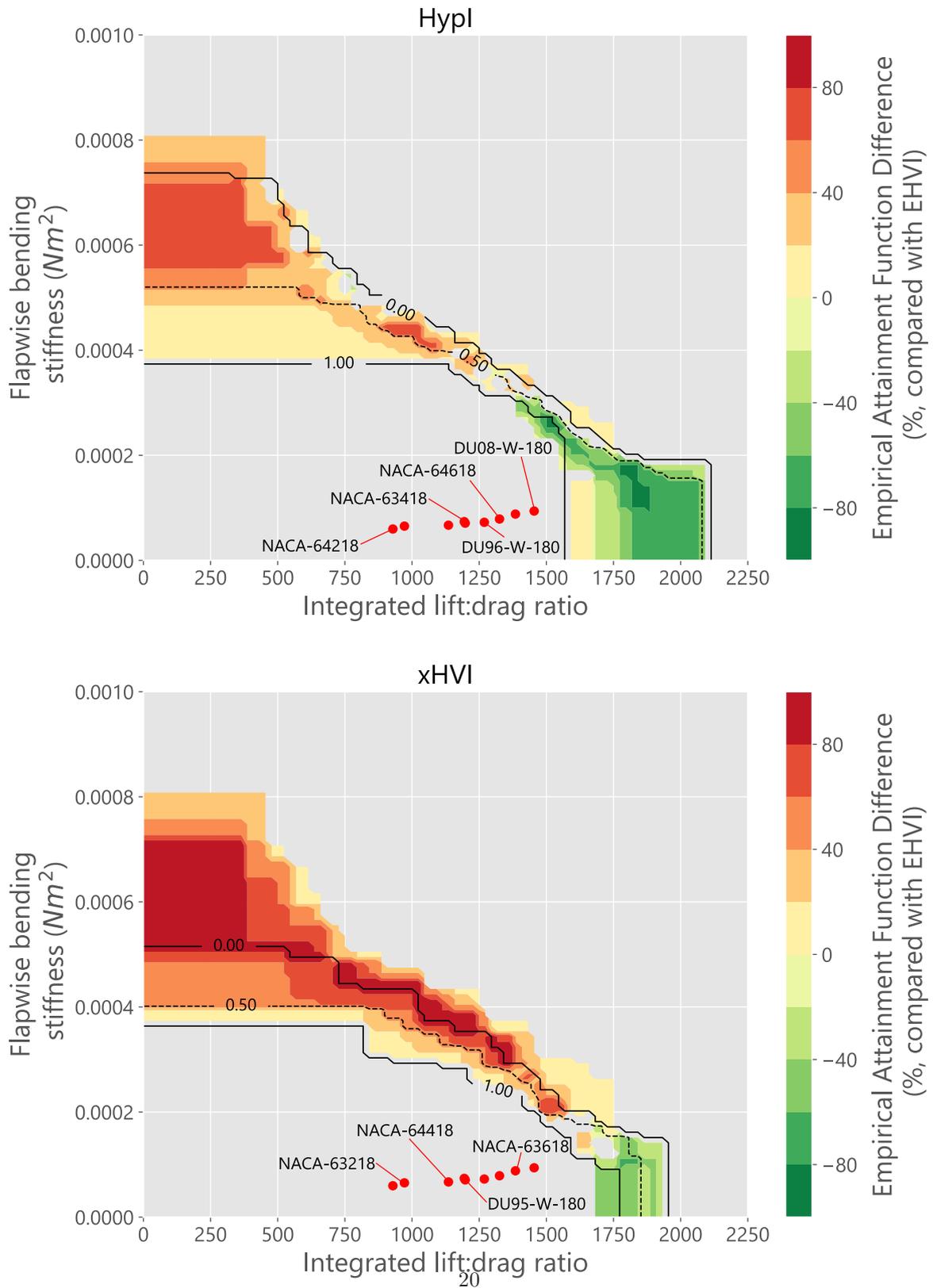}
 \caption{Comparison of results from Bayesian Optimisation of aerofoil shape after 800 evaluations, repeated 5 times using different random seeds. The black lines show the Empirical attainment functions for HypI (top) and xHVI (bottom); the colours indicate difference with EHVI. Performance results for standard aerofoils with the same thickness are also provided with the red dots.} \label{fig:aerofoil_pareto_front}
\end{figure}

Aerofoil shape optimisation was performed using the three BO approaches described earlier, given a budget of 800 evaluations, each repeated 5 times with different random seeds to test consistency. For comparison, six NACA and three DU aerofoils with the same thickness of 18\% were also evaluated~\cite{Timmer2013} (although DU08-W-180 dominates all other aerofoils). Results are shown in Fig.~\ref{fig:aerofoil_pareto_front}.

\begin{figure}[H]
 \includegraphics[width=\textwidth]{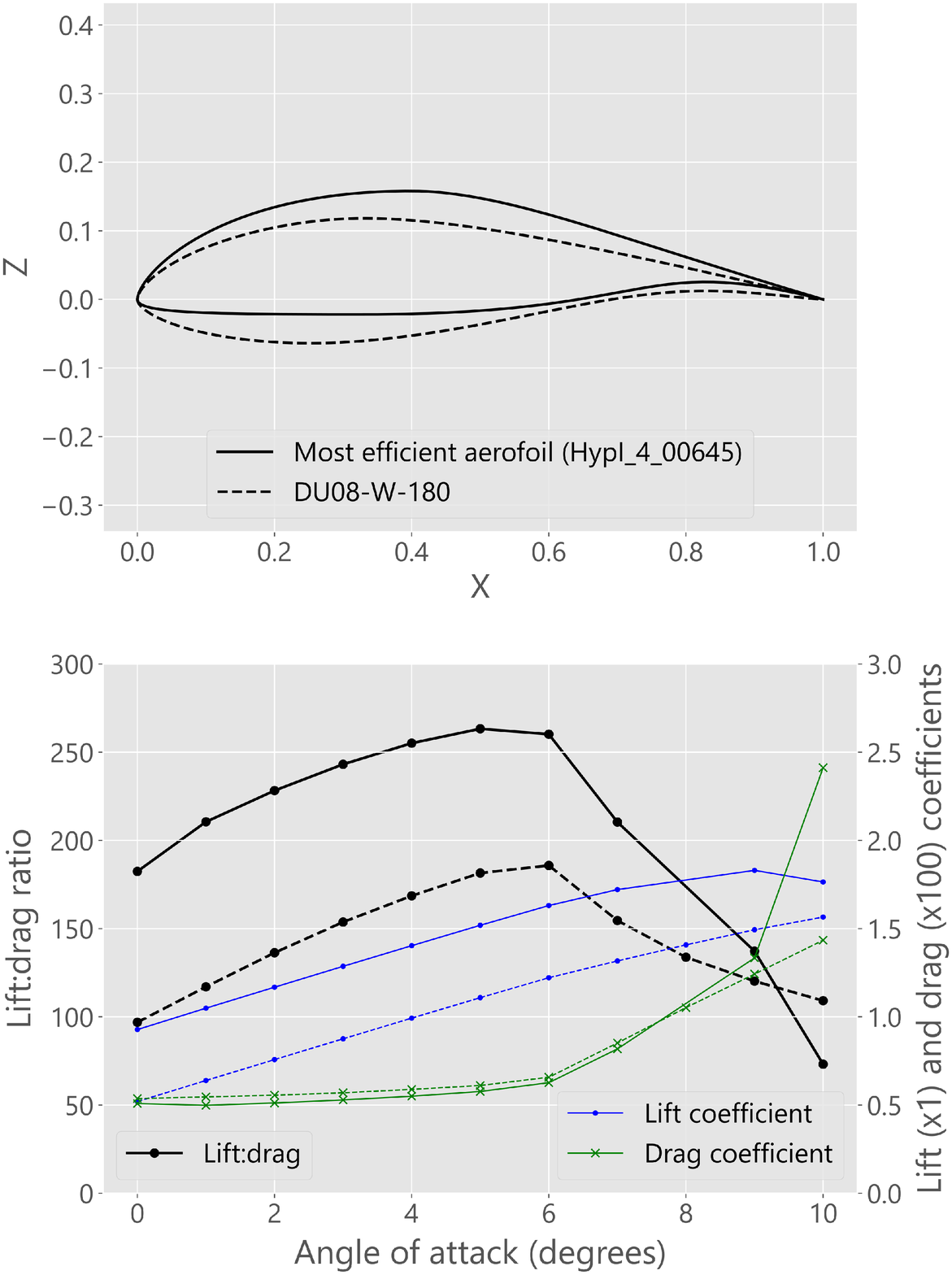}
 \caption{Comparison of the aerofoil with highest aerodynamic performance from the BO runs using HypI, with the standard aerofoil DU08-W-180.} \label{fig:best_aerodynamic_aerofoil}
\end{figure}

\begin{figure}[H]
 \includegraphics[width=\textwidth]{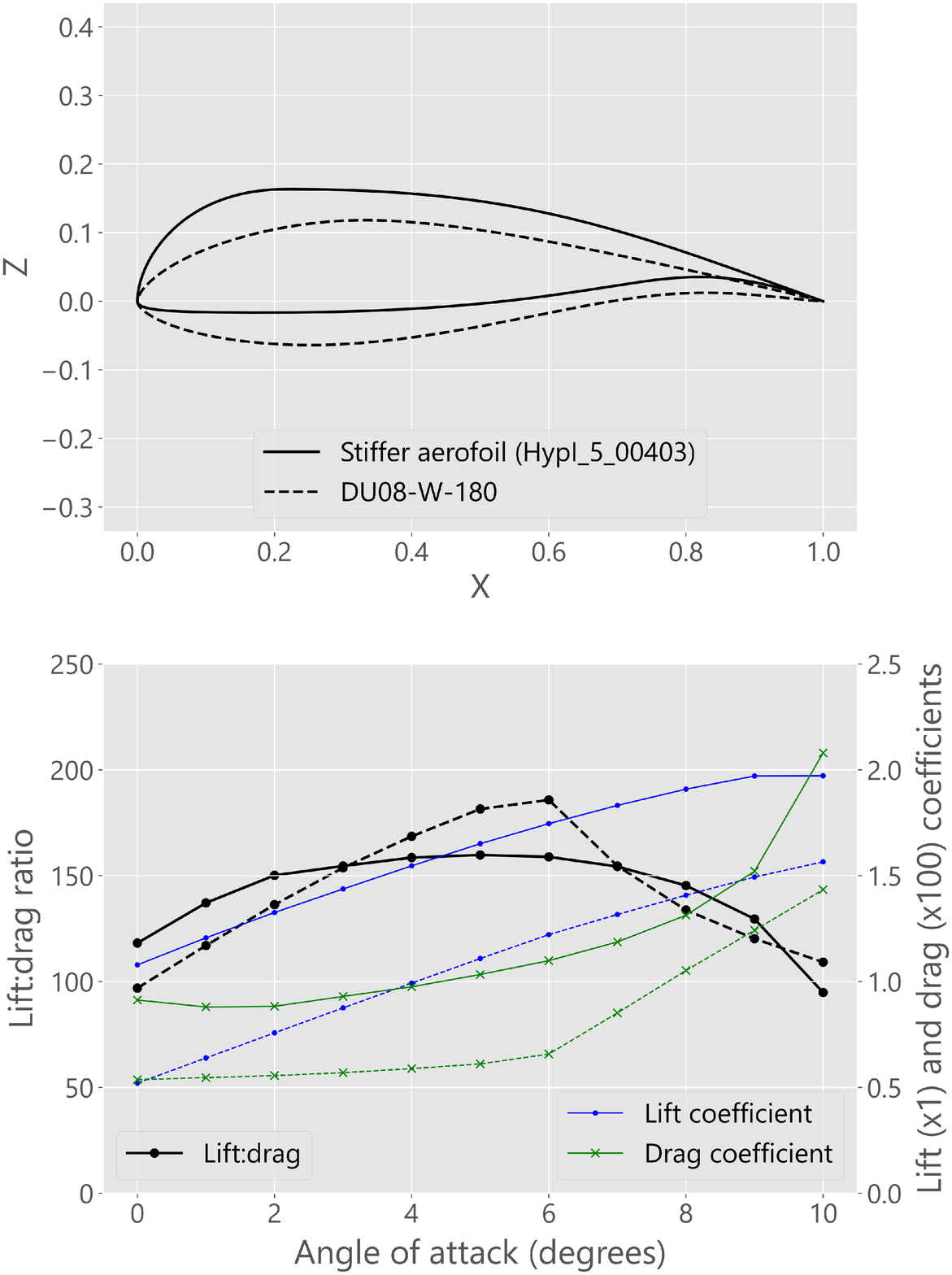}
 \caption{Comparison of the aerofoil found from the BO runs using HypI with the highest stiffness and equivalent aerodynamic performance to the standard aerofoil DU08-W-180.} \label{fig:stiff_aerodynamic_aerofoil}
\end{figure}

The most aerodynamically-efficient aerofoil found by HypI is shown in Fig.~\ref{fig:best_aerodynamic_aerofoil} and compared with DU08-W-180. The decision-maker will be interested in understanding the trade-off between the two objectives, thus Fig.~\ref{fig:stiff_aerodynamic_aerofoil} is included to present another aerofoil from the HypI optimisations, with almost identical aerodynamic performance to DU08-W-180, but 3.4 times the stiffness.

The main conclusions are:
\begin{itemize}
    \item The new parameterisation proposed here allows for \textit{a priori} constraints to be set on the desired aerofoil shape, largely avoiding unphysical designs.
    \item Standard aerofoil designs appear to have much less focus on structural stiffness than on aerodynamic performance.
    \item All three Bayesian Optimisation methods far outperform the standard aerofoils on these objectives.
    \item The multi-surrogate EHVI obtains stiffer aerofoils than either mono-surrogate, however, HypI obtains greater aerodynamic performance. (Not shown: EHVI's results also come at the expense of far longer runtimes, as previously illustrated in~\ref{fig:comp-time}).
    \item Of the two scalarising functions, HypI outperforms xHVI. Compared with EHVI, HypI does exhibit some variability in outcome with initial conditions. This effect should reduce with a higher evaluation budget.
\end{itemize}

\section{Conclusions and Future Work}\label{sec:conclusions}
This study has shown that careful attention should be paid in Bayesian optimisation (BO) to the `surprise' of objective values, relative to the mean and variance of the Gaussian process surrogate. This has an important effect on the exploration/exploitation trade-off. To assist with explicit management of this, a new exploration parameter based on normalisation of the data has been developed.

Taking this into account, an experimental investigation has then been performed on the effectiveness of two rather similar scalarising functions based on hypervolume (using the Expected Improvement acquisition function) and the multi-surrogate Expected Hypervolume Improvement acquisition function. The recently-introduced HypI outperforms the novel xHVI on every function.
Further investigation identified that visualising the infill criterion in the objective space is not sufficient to understand its effectiveness. Consideration should rather be given to how well its value can be learned by the Gaussian process surrogate, and in particular whether the resulting surface will have a shallow and consistent gradient over both non-dominated and dominated solutions. This is desirable, to avoid areas of the search space remaining unexplored due to their extreme unattractiveness to the Expected Improvement acquisition function.

The practical usefulness of mono-objective infill criteria consists largely in speed and in re-using existing methods from single objective BO. While the xHVI infill criterion has been applied in parallel to a real wind turbine control problem~\cite{Yu2020}, this paper applies xHVI, HypI and EHVI to the aerofoil design problem for wind turbine blades. A novel parameterisation for this task is introduced and shown to be valuable for \textit{a priori} control over the decision space. HypI is demonstrated to be a powerful tool for this problem, obtaining a very large number of aerofoils which dominate standard aerofoils, using only 800 total evaluations. 

There are two immediate areas for further investigation. First, this study has been performed using test functions without (aleatoric) evaluation uncertainty. It is important to now examine the influence of uncertainty on surrogate modelling of scalarising functions, since a Gaussian likelihood may not be a good assumption. Second, correlated objectives should be investigated, particularly as they have received as yet little attention in the multi-objective BO literature.

\bibliographystyle{splncs04}
\bibliography{main}
\end{document}